\documentclass[conference]{IEEEtran}

\def\BibTeX{{\rm B\kern-.05em{\sc i\kern-.025em b}\kern-.08em
    T\kern-.1667em\lower.7ex\hbox{E}\kern-.125emX}}

\usepackage{amsmath,amsfonts}
\usepackage{algorithm}
\usepackage{algorithmic}
\usepackage{multirow}
\usepackage{graphicx}
\usepackage{textcomp}
\usepackage{url}
\usepackage{hyperref}
\usepackage{xcolor}
\usepackage{colortbl}
\usepackage{tikz}
\usepackage{listings}
\usepackage{ctable}
\usepackage{subfigure}
\usepackage[most]{tcolorbox}
\usepackage{xspace}
\definecolor{dkgreen}{RGB}{0,64,0}
\definecolor{ltgray}{RGB}{245,245,245}
\definecolor{mauve}{RGB}{139,0,139}

\definecolor{myblue}{RGB}{53, 121, 139}
\definecolor{mygray}{RGB}{240, 255, 240}

\newcounter{obsnum}

\newcommand{\tweakedsim}{\raise.17ex\hbox{$\scriptstyle\mathtt{\sim}$}}
\newcommand{\plexusname}{ScaleGNN\xspace}

\begin{document}

\title{Communication-free Sampling and 4D Hybrid Parallelism for Scalable Mini-batch GNN Training}

\author{
\IEEEauthorblockN{
Cunyang Wei\IEEEauthorrefmark{1}, Siddharth Singh\IEEEauthorrefmark{2}, Aishwarya Sarkar\IEEEauthorrefmark{3}, Daniel Nichols\IEEEauthorrefmark{4},
Tisha Patel\IEEEauthorrefmark{1}, \\Aditya K.~Ranjan\IEEEauthorrefmark{5}, Sayan Ghosh\IEEEauthorrefmark{6}, Ali Jannesari\IEEEauthorrefmark{3}, Nathan R. Tallent\IEEEauthorrefmark{6}, Abhinav Bhatele\IEEEauthorrefmark{1}
}
\IEEEauthorblockA{
\IEEEauthorrefmark{1}\itshape University of Maryland, College Park, USA\\
\IEEEauthorrefmark{2}\itshape NVIDIA, Inc. \\
\IEEEauthorrefmark{3}\itshape Iowa State University, USA\\
\IEEEauthorrefmark{4}\itshape Lawrence Livermore National Laboratory, USA\\
\IEEEauthorrefmark{5}\itshape Google, Inc. \\
\IEEEauthorrefmark{6}\itshape Pacific Northwest National Laboratory, USA\\
E-mail: cunyang@umd.edu, sidsingh@nvidia.com, danielnichols@llnl.gov, tpatel21@terpmail.umd.edu,\\
\{asarkar1,jannesari\}@iastate.edu,
adikranjan@google.com,
\{sayan.ghosh,tallent\}@pnnl.gov, bhatele@cs.umd.edu
}
}

\maketitle

\begin{abstract}
Graph neural networks (GNNs) are widely used for learning on graph datasets
derived from various real-world scenarios.  Learning from extremely large
graphs requires distributed training, and mini-batching with sampling is a
popular approach for parallelizing GNN training.  Existing distributed
mini-batch approaches have significant performance bottlenecks due to expensive
sampling methods and limited scaling when using data parallelism.  In this
work, we present \plexusname, a 4D parallel framework for scalable mini-batch
GNN training that combines communication-free distributed sampling, 3D parallel
matrix multiplication (PMM), and data parallelism. \plexusname introduces a
uniform vertex sampling algorithm, enabling each process (GPU device) to
construct its local mini-batch, i.e., subgraph partitions without any
inter-process communication.  3D PMM enables scaling mini-batch training to
much larger GPU counts than vanilla data parallelism with significantly lower
communication overheads.
We also present additional optimizations to overlap sampling with training,
reduce communication overhead by sending data in lower precision, kernel
fusion, and communication-computation overlap.
We evaluate \plexusname on five graph datasets and demonstrate strong scaling
up to 2048 GPUs on Perlmutter, 2048 GCDs on Frontier, and 1024 GPUs on Tuolumne.  On Perlmutter,
\plexusname achieves $3.5\times$ end-to-end training speedup over the
SOTA baseline on ogbn-products.

\end{abstract}

\section{Introduction}
\label{sec:intro}
Graph neural networks or GNNs~\cite{GNN-model} are becoming increasingly
popular for learning from graph datasets found in the real world around us.
They power tasks such as recommendation systems~\cite{GNN-recommender,
GNN-social-recommendation}, fraud detection~\cite{GNN-fraud}, and scientific
discovery~\cite{GNN-nature}. Most modern GNNs follow the message-passing
pattern~\cite{GNN-message-passing}: at each layer, a vertex aggregates
information from its neighbors and then updates its embedding. Graph
Convolutional Networks or GCNs~\cite{KipfW16} are a canonical and widely used
instantiation of this pattern.
 
GNN training has two paradigms: full-graph training and mini-batch training.
Full-graph training processes all vertices in each iteration, offering a
regular execution structure but quickly hitting memory and communication
bottlenecks as graphs grow. Mini-batch training instead operates on sampled
subgraphs, reducing the working set to fit within GPU memory. Common sampling
strategies include node-wise neighbor sampling~\cite{graphsage}, layer-wise
sampling~\cite{fastgcn, ladies}, and subgraph-based
sampling~\cite{cluster-gcn,graphsaint}. Recent studies have demonstrated that
mini-batch training can converge faster and reach higher accuracy than
full-graph training~\cite{bajaj2024graph}, making it increasingly preferred in
practice.

GNN training workloads are irregular, memory-bound, and tightly coupled to
graph structure, which makes them difficult to run efficiently on large GPU
systems. Parallelization of full-graph GNN training typically requires 1D--3D
algorithms to parallelize sparse and dense matrix operations, but full-graph
iterations remain expensive on large graphs~\cite{cagnet,
cagnet-sparsity-aware, ranjan:sc2025}. Parallel approaches for mini-batch
training use vanilla data parallelism to assign mini-batches to workers (GPUs),
and use neighbor sampling with remote feature fetching. However, their reliance
on CPU-based sampling and cross-device feature access limits
scalability~\cite{distdgl, kaler2023communication, massivegnn,
tripathy2024distributed}. Hence, despite previous efforts, distributed
mini-batch GNN training still struggles to scale efficiently on large GPU
systems.

We identify two key limitations. First, neighbor sampling pipelines often
incur high sampling costs that can erase the benefits of mini-batching. In many
frameworks, sampling is executed on CPUs and requires frequent communication for
neighbor and feature access, making sampling a critical performance bottleneck.
Second, most distributed mini-batch frameworks rely on vanilla data parallelism for scaling to multiple GPUs.
While data parallelism can
improve throughput, our experiments show that it does not necessarily reduce
end-to-end training time (Section~\ref{subsec:e2e-perf}). As a result, there is a need for a parallel GNN training framework
that can sample efficiently on GPUs, preserve model accuracy, and scale to
the largest HPC platforms.

In this work, we introduce \plexusname, an open-source, highly scalable, 4D parallel
framework for mini-batch GNN training. To address the sampling bottleneck, \plexusname
uses a uniform vertex sampling algorithm that requires no inter-device
communication. Every process constructs its local mini-batch subgraph partitions
from local data alone.  We further overlap sampling with training through a
pipelined execution schedule, effectively removing sampling from the critical
path.

For distributed training, \plexusname organizes GPUs into a 4D virtual grid,
and combines data parallelism with 3D parallel matrix multiplication (PMM).
Data parallelism enables each DP group to process independent mini-batches with
gradient synchronization via all-reduce.  3D
PMM~\cite{agarwal-3d,ranjan:sc2025,singh:sc2024,singh:ipdps2022} is used to
parallelize sparse and dense matrix multiplication operations of each GNN layer
across a 3D virtual grid of GPUs.  We further optimize \plexusname by incorporating low-precision
collective communication, kernel fusion, and communication--computation overlap
to reduce epoch time. Our design builds on and extends recent progress in 3D
parallelism for full-graph GNN training~\cite{cagnet,ranjan:sc2025} and for
large-scale deep learning~\cite{singh:ipdps2022, colossalai2023unified,oslo}.

We summarize our key contributions as follows:
\begin{itemize}
  \item We design and implement \plexusname, an open source 4D parallel mini-batch 
  GNN training framework that combines distributed sampling, 3D PMM,
  and data parallel training. To the best of our knowledge,
  \plexusname is the first GNN framework to combine data parallelism,
  3D PMM, and distributed sampling in a unified framework.
  \item We propose a communication-free distributed sampling algorithm based on
  uniform vertex sampling. Our sampling strategy reaches 81.3\% test accuracy on 
  ogbn-products, outperforming both GraphSAINT~\cite{graphsaint} and GraphSAGE~\cite{graphsage}.
  \item We identify several optimizations, including fully overlapping sampling 
  with training, low-precision collective communication, kernel fusion, and 
  communication-computation overlap.
  \item We evaluate \plexusname on five graph datasets and demonstrate strong 
  scaling up to 2048 GPUs on Perlmutter, 2048 GCDs on Frontier, and 1024 GPUs on Tuolumne. On Perlmutter, \plexusname achieves $3.5\times$ end-to-end training
  speedup over the state-of-the-art baseline on ogbn-products, while matching or
  exceeding their accuracy.
\end{itemize}

\section{Background and Related Work}
\label{sec:bg}
This section reviews graph neural networks and mini-batch sampling strategies,
then introduces distributed GNN training systems and identifies gaps that 
motivate our work.

\subsection{Graph Neural Networks}

Graph neural networks (GNNs) learn vertex representations by repeatedly 
aggregating information from neighboring vertices~\cite{GNN-model, 
gnn-acceleration-survey, luo2024classic, wu2020comprehensive, liu2022survey}. 
In a typical message-passing GNN with $L$ layers, each layer $l$ updates the 
embedding of every vertex $v$ in two steps. First, an \textit{aggregation} 
step collects embeddings from all neighbors $u \in \mathcal{N}(v)$:
\begin{equation}
  \mathbf{a}_v^{(l)} = \textsc{Aggregate}\!\left(\left\{\mathbf{h}_u^{(l-1)} : u \in \mathcal{N}(v)\right\}\right),
\end{equation}
where $\mathbf{h}_u^{(l-1)}$ is the embedding of vertex $u$ from the previous 
layer. Then, an \textit{update} step combines the aggregated message with the 
vertex embedding through a learnable transformation:
\begin{equation}
  \mathbf{h}_v^{(l)} = \textsc{Update}\!\left(\mathbf{h}_v^{(l-1)},\;\mathbf{a}_v^{(l)}\right).
\end{equation}
The input vertex features serve as the initial embeddings $\mathbf{h}_v^{(0)}$. 
After $L$ layers, each vertex embedding captures structural and feature 
information from its $L$-hop neighborhood.

Graph Convolutional Networks (GCNs)~\cite{KipfW16} instantiate this framework 
with a specific choice: aggregation computes a normalized sum over neighbor 
embeddings, and update applies a shared weight matrix followed by a 
nonlinearity. Concretely, a GCN layer computes
\begin{equation}
  \mathbf{H}^{(l)} = \sigma\!\left(\hat{\mathbf{D}}^{-\frac{1}{2}}\hat{\mathbf{A}}\,\hat{\mathbf{D}}^{-\frac{1}{2}}\,\mathbf{H}^{(l-1)}\mathbf{W}^{(l)}\right),
\end{equation}
where $\hat{\mathbf{A}} = \mathbf{A} + \mathbf{I}$ is the adjacency matrix with 
added self-loops, $\hat{\mathbf{D}}$ is the corresponding degree matrix, $\mathbf
{W}^{(l)}$ is a trainable weight matrix, and $\sigma$ is a nonlinear activation. 
In matrix form, the forward pass of a GCN layer consists of sparse matrix 
multiplication (SpMM) for aggregation, followed by dense matrix multiplication 
(GEMM) for feature transformation. 

\subsection{Mini-batch GNN Training}

Full-graph training updates embeddings for all vertices in every iteration. 
This approach is simple, but it quickly becomes memory- and communication-intensive at scale.

Mini-batch training instead optimizes the same objective using stochastic 
gradients computed from a small set of target vertices and their sampled 
$L$-hop neighborhoods. Recent papers~\cite{bajaj2024graph} also suggest 
that mini-batch training can converge faster and achieve higher accuracy 
than full-graph training, which further motivates scalable mini-batch GNN 
systems.

\begin{figure}[h]
  \centering
  \includegraphics[width=\columnwidth]{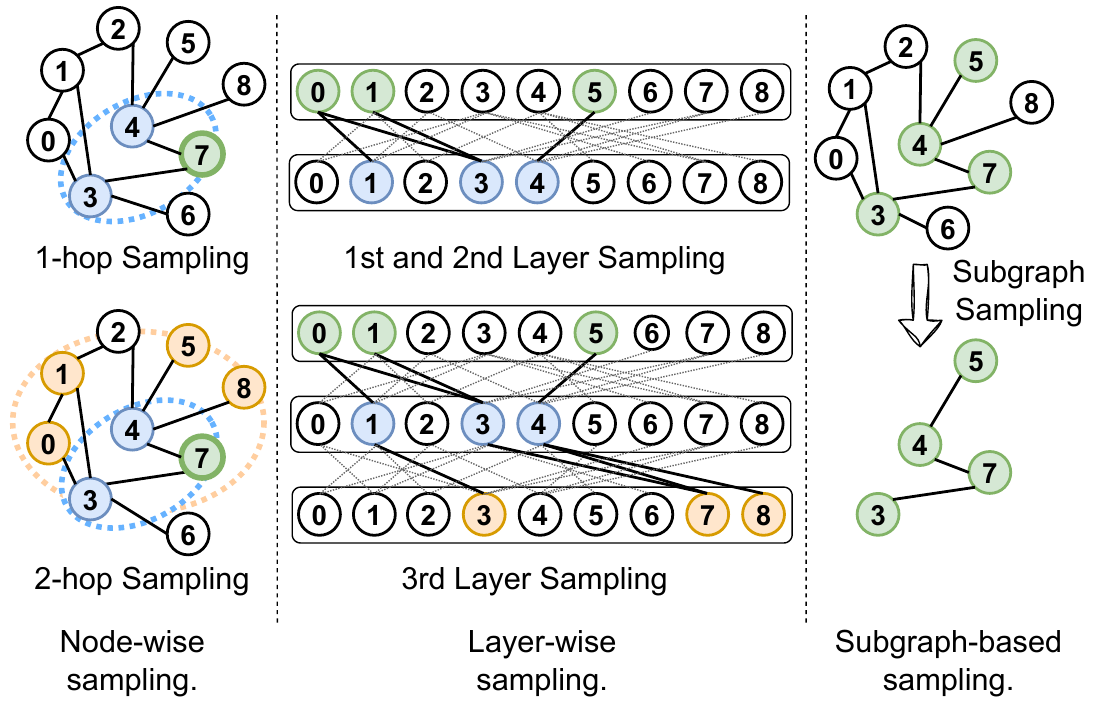}
  \caption{Three families of sampling algorithms. (a) Node-wise sampling. 
  (b) Layer-wise sampling. (c) Subgraph-based sampling.}
  \label{fig:bg-algorithms}
\end{figure}

As shown in Figure~\ref{fig:bg-algorithms}, three families of sampling 
algorithms are widely used:

\noindent\textbf{Node-wise sampling}~\cite{graphsage, dai2018learning} 
independently samples up to $k_l$ neighbors per vertex at each layer, as 
popularized by GraphSAGE~\cite{graphsage}. Its simplicity and accuracy make it 
common in practice, although it can suffer from 
\textit{neighborhood explosion} as fan-out grows with depth.

\noindent\textbf{Layer-wise sampling}~\cite{fastgcn, zou2019layer, 
huang2018adaptive,ladies} bounds the number of sampled vertices per layer. 
It avoids neighborhood explosion, but it can miss informative neighbors and 
increase gradient variance.

\noindent\textbf{Subgraph-based sampling}~\cite{zeng2019accurate, cluster-gcn, 
graphsaint, bai2021ripple} trains on connected subgraphs sampled from the 
original graph. Cluster-GCN~\cite{cluster-gcn} uses graph clustering, while 
GraphSAINT~\cite{graphsaint} uses random walk or edge sampling with 
normalization for bias correction. These methods offer good locality, but 
their performance depends on subgraph quality.

\begin{figure*}[t]
  \centering
  \includegraphics[width=1\textwidth]{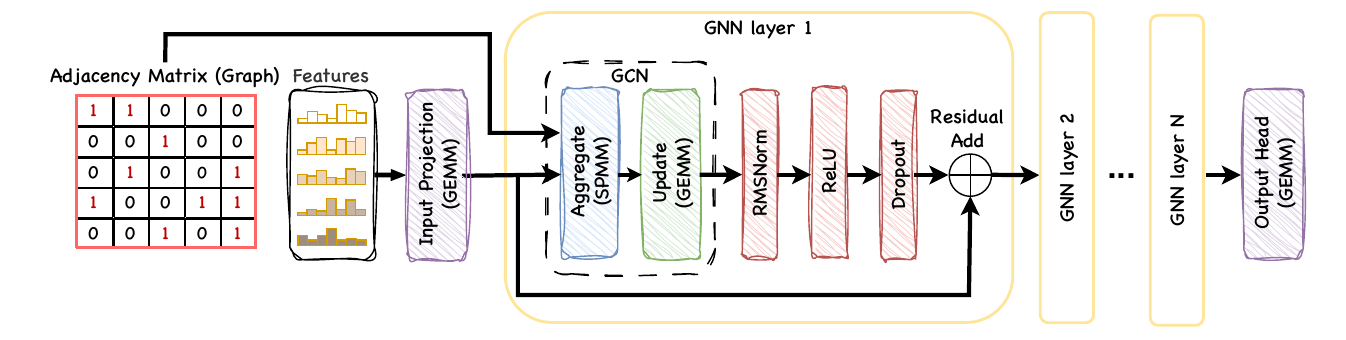}
  \caption{Model architecture in \plexusname. Vertex features and the graph
  adjacency matrix enter an input projection (GEMM) that maps features to a
  uniform hidden dimension. The projected features then pass through
  GNN layers, each comprising a GCN convolution (SpMM aggregation followed by
  GEMM update), RMSNorm, ReLU, dropout, and a residual connection. An output head (GEMM) produces the final 
  class logits.}
  \label{fig:overview}
\end{figure*}

\subsection{Distributed GNN Training Frameworks}

A number of systems~\cite{besta2024parallel, shao2024distributed,
lin2023comprehensive, pipegcn} scale GNN training across multiple GPUs or
machines.

\vspace{0.08in}
\noindent\textbf{Full graph systems.}
NeuGraph~\cite{neugraph} and ROC~\cite{roc-gnn} are among the first systems to 
distribute full-graph GNN training across GPUs. They partition the graph and 
schedule computation to reduce cross-device data movement. 
CAGNET~\cite{cagnet,cagnet-sparsity-aware} formulates each GNN layer as sparse 
and dense matrix operations and applies 1D, 1.5D, 2D, and 3D parallel 
algorithms drawn from distributed linear algebra. BNS-GCN~\cite{bns-gcn} 
reduces communication in full-graph training by sampling boundary vertices.
Plexus~\cite{ranjan:sc2025} is a full-graph GNN training system that uses 3D 
parallelism to distribute the workload across GPUs.
GNNPipe~\cite{chen2023gnnpipe} and Mithril~\cite{chen2025mithril} explore
pipelined layer-level model parallelism, partitioning GNN layers across GPUs
to reduce communication volume.
These systems handle large-scale graphs effectively, but full-graph training
remains expensive on very large graphs.

\vspace{0.08in}
\noindent\textbf{Mini-batch systems.}
DistDGL~\cite{distdgl} extends the Deep Graph Library~\cite{dgl} to multiple 
machines. It partitions the graph and stores vertex features in a distributed 
key-value store. Each worker runs sampling and training locally but must fetch 
remote features through the network, which can become a bottleneck at scale. 
MassiveGNN~\cite{massivegnn} builds on DistDGL with optimized feature fetching 
and supports training on graphs with billions of edges. 
SALIENT++~\cite{kaler2023communication} improves CPU-based sampling throughput 
and caches frequently accessed features to reduce remote feature fetches. 
Tripathy et al.~\cite{tripathy2024distributed} extend the CAGNET~\cite{cagnet} 
approach to mini-batch training by using distributed SpGEMM to parallelize 
the neighbor sampling process.
BGL~\cite{liu2023bgl} and FastGL~\cite{zhu2024fastgl} optimize GPU-side
sampling and data I/O to reduce preprocessing overhead, while
GSplit~\cite{polisetty2023gsplit} introduces split-parallelism to eliminate
redundant sampling across GPUs. However, none of these systems combines tensor 
parallelism with data parallelism for mini-batch GNN training or explores 
efficient communication-free sampling algorithms. \plexusname fills this gap.

\section{GNN Model and Sampling Strategy}
\label{sec:overview}
This section describes the GNN architecture used in this paper and the 
operator-level forward and backward passes that underpin our 4D parallel 
training design in Section~\ref{sec:training}.

\subsection{Architecture Overview}

We build on GCN~\cite{KipfW16}, one of the most widely adopted 
message-passing GNNs. 
Following recent findings that normalization, dropout, and residual connections
substantially improve GNN accuracy on node classification
benchmarks~\cite{luo2024classic}, we augment each GNN layer with these
components as shown in Figure~\ref{fig:overview}.
The input projection first maps raw vertex features to a uniform hidden 
dimension $d_h$, enabling residual connections across all layers.
The projected features then pass through $L$ stacked GNN layers, each 
consisting of a GCN convolution (SpMM + GEMM), RMS normalization, ReLU 
activation, dropout, and a residual connection. The output head projects the 
final hidden representation to class logits and computes the loss. Each 
component can be enabled or disabled without changing the parallelization strategy.

\subsection{Forward Pass}

We now walk through the detailed computation of each stage and explain the 
role of every operator.

\subsubsection{Input Projection}
The input projection maps raw vertex features 
$X_{\mathrm{in}} \in \mathbb{R}^{N \times d_{\mathrm{in}}}$ to hidden dimension $d_h$:
\begin{equation}
    X_{h,0} = X_{\mathrm{in}} \, W_{\mathrm{in}} \quad \text{(GEMM)}
\end{equation}
where $N$ is the number of vertices and $W_{\mathrm{in}} \in \mathbb{R}^{d_{\mathrm{in}} \times d_h}$.

\subsubsection{GNN Layers ($l \in \{1, \dots, L\}$)}
Each GNN layer applies the following sequence of operators:

\noindent\textbf{GCN convolution} has two steps: sparse neighborhood aggregation 
(SpMM) with normalized adjacency $A = \hat{D}^{-\frac{1}{2}}\hat{A}\,\hat{D}^{-\frac{1}{2}}$~\cite{KipfW16}, 
followed by dense feature transformation (GEMM) with a learned weight matrix.
This mix of sparse and dense computation characterizes GNN workloads and 
drives the parallelization strategies developed in Section~\ref{sec:training}.
\begin{align}
    H_{\mathrm{agg},l} &= A \, X_{h,l-1} \quad \text{(SpMM)} \label{eq:spmm-fwd} \\
    X_{\mathrm{conv},l} &= H_{\mathrm{agg},l} \, W_l \quad \text{(GEMM)} \label{eq:gemm-fwd} 
\end{align}

\noindent\textbf{RMS normalization}~\cite{rmsnorm} (Eq.~\ref{eq:rms-norm}) 
rescales each feature vector by its root mean square.
Compared with layer normalization~\cite{layer-norm}, it omits mean centering 
and reduces per-vertex computation while preserving training stability.

\noindent\textbf{ReLU}~\cite{xu2015empirical} (Eq.~\ref{eq:relu}) applies an
element-wise nonlinearity that enables the network to learn non-linear vertex
representations.

\noindent\textbf{Dropout}~\cite{srivastava2013improving} (Eq.~\ref{eq:dropout}) 
randomly zeros a fraction of activations during training to reduce overfitting.

\noindent\textbf{Residual connections}~\cite{resnetCVPR}
(Eq.~\ref{eq:residual-add}) add each layer's input to its output, mitigating
over-smoothing~\cite{deeper-insights-gnn} and improving gradient flow.

\begin{align}
    X_{n,l} &= \text{RMSNorm}(X_{\mathrm{conv},l}) \label{eq:rms-norm} \\
    X_{r,l} &= \text{ReLU}(X_{n,l}) \label{eq:relu} \\
    X_{d,l} &= X_{r,l} \odot M_{\mathrm{drop},l} \quad \text{(Dropout)} \label{eq:dropout} \\
    X_{h,l} &= X_{d,l} + X_{h,l-1} \quad \text{(Residual Add)} \label{eq:residual-add}
\end{align}

\subsubsection{Output Head and Loss}
The final hidden representation is projected to output logits, then compared
with labels $Y$:
\begin{align}
    O &= X_{h,L} \, W_{\mathrm{out}} \quad \text{(GEMM)} \\
    \mathcal{L} &= \text{Loss}(O,\, Y)
\end{align}
where $W_{\mathrm{out}} \in \mathbb{R}^{d_h \times d_{\mathrm{out}}}$ and 
$d_{\mathrm{out}}$ is the number of classes.
We use cross-entropy for single-label classification and binary cross-entropy 
for multi-label tasks.

\subsection{Backward Pass}

The backward pass reverses the forward pass and propagates 
$\nabla_{O}$ through all operators.

\subsubsection{Output Head Backward}
\begin{align}
    \nabla_{W_{\mathrm{out}}} &= X_{h,L}^T \, \nabla_{O} \quad \text{(GEMM)} \\
    \nabla_{X_{h,L}} &= \nabla_{O} \, W_{\mathrm{out}}^T \quad \text{(GEMM)}
\end{align}

\subsubsection{GNN Layer Backward (for each $l = L, \dots, 1$)}
The gradient $\nabla_{X_{h,l}}$ is split identically into the main branch and 
residual skip path. After element-wise backward operations (dropout mask, ReLU 
mask, RMSNorm backward), the main branch yields $\nabla_{X_{\mathrm{conv},l}}$.
The SpMM and GEMM gradients are:
\begin{align}
    \nabla_{W_l} &= H_{\mathrm{agg},l}^T \, \nabla_{X_{\mathrm{conv},l}} \quad \text{(GEMM)} \label{eq:gemm-bwd-w} \\
    \nabla_{H_{\mathrm{agg},l}} &= \nabla_{X_{\mathrm{conv},l}} \, W_l^T \quad \text{(GEMM)} \label{eq:gemm-bwd-h} \\
    \nabla_{X_{h,l-1}^{\mathrm{conv}}} &= A^T \nabla_{H_{\mathrm{agg},l}} \quad \text{(SpMM)} \label{eq:spmm-bwd}
\end{align}
The final gradient merges both paths: $\nabla_{X_{h,l-1}} = \nabla_{X_{h,l-1}^{\mathrm{conv}}} + \nabla_{X_{h,l-1}^{\mathrm{skip}}}$.

\subsubsection{Input Projection Backward}
\begin{align}
    \nabla_{W_{\mathrm{in}}} &= X_{\mathrm{in}}^T \, \nabla_{X_{h,0}} \quad \text{(GEMM)} \\
    \nabla_{X_{\mathrm{in}}} &= \nabla_{X_{h,0}} \, W_{\mathrm{in}}^T \quad \text{(GEMM)}
\end{align}

\begin{figure*}[t]
    \centering
    \includegraphics[width=1\textwidth]{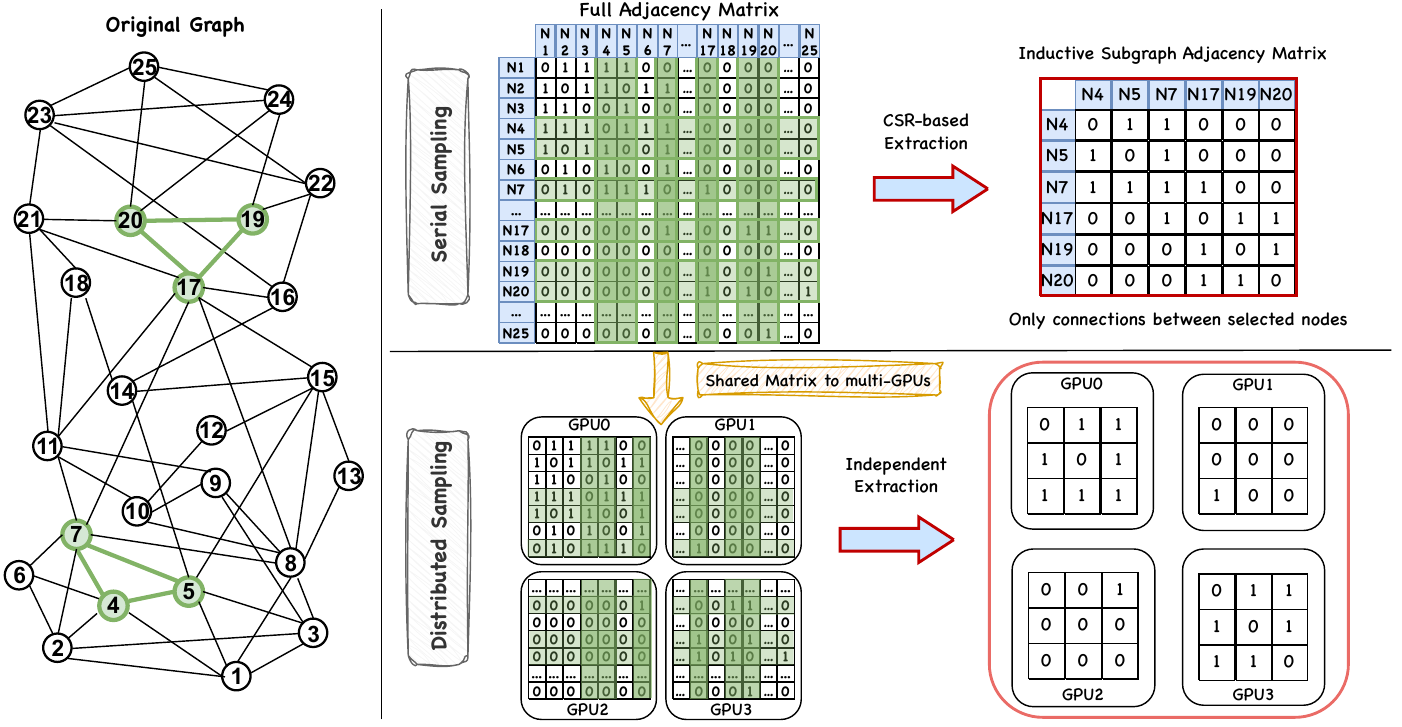}
    \caption{\plexusname uniform vertex sampling. (Left) Uniform vertex sampling
    on the original graph. Selected vertices are shown in green. (Upper right)
    The full adjacency matrix with sampled rows and columns highlighted, and the
    induced subgraph adjacency retaining only edges between selected vertices.
    (Lower right) Distributed sampling: the adjacency matrix is partitioned
    across GPUs, each independently sampling its local shard.}
    \label{fig:sampling}
\end{figure*}

\subsection{Uniform Vertex Sampling}
\label{subsec:sampling-overview}

Mini-batch training reduces cost by operating on a sampled subgraph.
When scaled to multi-GPU systems, however, the two most widely adopted
sampling algorithms both require cross-device communication:
GraphSAGE~\cite{graphsage} must fetch multi-hop neighbors from remote GPUs to
construct each vertex's receptive field, while
GraphSAINT~\cite{graphsaint} requires normalization coefficients derived from
global graph structure. As a result, distributed frameworks built on these sampling
algorithms must perform extensive cross-device communication during the
sampling phase, making sampling a critical bottleneck in distributed
mini-batch GNN training.
We design \emph{uniform vertex sampling} with the explicit goal of eliminating
all cross-device communication during sampling while preserving model accuracy.
We describe the algorithm in detail below.

\subsubsection{Vertex Sampling}
At each training step, we sample a subset $\mathcal{S} \subset V$ of $B$ vertices 
uniformly without replacement:
\begin{equation}
    \mathcal{S} \sim \text{Uniform}\!\left(\binom{V}{B}\right), \quad |\mathcal{S}| = B.
\label{eq:node-sample}
\end{equation}
Every vertex $v \in V$ has inclusion probability $\Pr[v \in \mathcal{S}] = B/N$.
The sampled set $\mathcal{S}$ is used as both target vertices (for predictions 
and loss) and source vertices (for aggregation).

\subsubsection{Induced Subgraph Construction}
Given $\mathcal{S}$, we construct the vertex-induced subgraph 
$G_{\mathcal{S}} = (\mathcal{S}, E_{\mathcal{S}})$ with edges 
whose endpoints are both in $\mathcal{S}$:
\begin{equation}
    E_{\mathcal{S}} = \{(u,v) \in E \mid u \in \mathcal{S} \wedge v \in \mathcal{S}\}.
\label{eq:induced-edges}
\end{equation}
The adjacency $A_{\mathcal{S}} \in \mathbb{R}^{B \times B}$ inherits normalized 
weights from $A$, and $A_{\mathcal{S}}^T$ is built alongside it for backward
SpMM (Eq.~\ref{eq:spmm-bwd}).
The same subgraph $G_{\mathcal{S}}$ is reused across all $L$ layers.

\subsubsection{Unbiased Edge Rescaling}
Using only the induced subgraph drops edges to neighbors outside $\mathcal{S}$, so 
mini-batch aggregation for sampled vertex $v \in \mathcal{S}$ is
\begin{equation}
    \tilde{h}_v = \sum_{u \in \mathcal{N}(v) \cap \mathcal{S}} a_{vu}\, x_u,
\label{eq:biased-agg}
\end{equation}
which underestimates full-graph aggregation $h_v = \sum_{u \in \mathcal{N}(v)} a_{vu}\, x_u$.
For sampled $v \in \mathcal{S}$ and neighbor $u \neq v$, the conditional 
inclusion probability is
\begin{equation}
    p = \Pr[u \in \mathcal{S} \mid v \in \mathcal{S}] = \frac{B-1}{N-1}.
\label{eq:edge-scale}
\end{equation}
We define rescaled adjacency
\begin{equation}
    \tilde{a}_{vu} =
    \begin{cases}
        a_{vu} / p & \text{if } u \neq v, \\
        a_{vv}     & \text{if } u = v,
    \end{cases}
\label{eq:rescaled-adj}
\end{equation}
leaving self-loops unchanged because $v$ is always present in its own sample.
Then mini-batch aggregation is an unbiased estimator of full-graph aggregation 
at each layer:
\begin{align}
    &\mathbb{E}_{\mathcal{S}}\!\left[\sum_{u \in \mathcal{N}(v) \cap \mathcal{S}} \tilde{a}_{vu}\, x_u \;\middle|\; v \in \mathcal{S}\right] \nonumber \\
    &\quad = a_{vv}\, x_v + \sum_{\substack{u \in \mathcal{N}(v),\, u \neq v}} a_{vu}\, x_u
    = h_v.
\label{eq:unbiased-proof}
\end{align}
This importance-sampling rescaling is a standard technique in subgraph-based 
GNN training. GraphSAINT~\cite{graphsaint} and BNS-GCN~\cite{bns-gcn} apply 
similar edge normalization to correct for sampling bias.
Crucially, our rescaling factor $p$ depends only on global constants $B$ and 
$N$, so each GPU can apply it independently without any communication.
Section~\ref{subsec:distributed-sampling} describes how multiple GPUs
collaboratively construct the distributed mini-batch subgraph.

\subsubsection{Feature and Label Slicing}
We extract features and labels for the sampled vertices:
\begin{equation}
    X_{\mathcal{S}} = X_{\mathrm{in}}[\mathcal{S}] \in \mathbb{R}^{B \times d_{\mathrm{in}}}, \quad
    Y_{\mathcal{S}} = Y[\mathcal{S}].
\label{eq:feature-slice}
\end{equation}
All intermediate activations therefore have row dimension $B$ instead of $N$, 
and the loss is computed only on sampled vertices.

\subsubsection{Mini-batch Training Step}
\label{subsec:minibatch-step}

\begin{algorithm}[h]
    \caption{Mini-batch GNN Training Step}\label{alg:minibatch}
    \begin{algorithmic}[1]
    \REQUIRE Graph $G=(V,E)$ with adjacency $A$, features $X_{\mathrm{in}}$, labels $Y$, batch size $B$
    \STATE $\mathcal{S} \sim \text{Uniform}\!\left(\binom{V}{B}\right)$ \COMMENT{Eq.~\ref{eq:node-sample}}
    \STATE $E_{\mathcal{S}} \leftarrow \{(u,v) \in E \mid u \in \mathcal{S} \wedge v \in \mathcal{S}\}$ \COMMENT{Eq.~\ref{eq:induced-edges}}
    \STATE $\tilde{A}_{\mathcal{S}} \leftarrow$ rescale $A_{\mathcal{S}}$ via Eq.~\ref{eq:rescaled-adj}
    \STATE $X_{\mathcal{S}} \leftarrow X_{\mathrm{in}}[\mathcal{S}]$; \quad $Y_{\mathcal{S}} \leftarrow Y[\mathcal{S}]$ \COMMENT{Eq.~\ref{eq:feature-slice}}
    \STATE $O \leftarrow \textsc{ForwardPass}(\tilde{A}_{\mathcal{S}},\, X_{\mathcal{S}})$ \COMMENT{Section~\ref{sec:overview}}
    \STATE $\mathcal{L} \leftarrow \textsc{Loss}(O,\, Y_{\mathcal{S}})$
    \STATE Backpropagate using $\tilde{A}_{\mathcal{S}}^T$; update parameters
    \end{algorithmic}
    \end{algorithm}

Algorithm~\ref{alg:minibatch} summarizes a complete mini-batch training step.
Each iteration uniformly samples $B$ vertices to form $\mathcal{S}$, extracts 
the vertex-induced subgraph, and rescales edge weights for unbiased aggregation
(Eq.~\ref{eq:unbiased-proof}).
All subsequent computation, including the forward pass, loss evaluation, 
backpropagation, and parameter updates, operates on a compact subgraph of $B$ 
vertices rather than the full graph of $N$ vertices.
Crucially, when distributed to multiple processes, every step in 
Algorithm~\ref{alg:minibatch} depends only on the local graph partition.
The entire sampling and subgraph construction procedure is therefore 
communication-free. We detail the distributed implementation in Section~\ref{sec:training}.

\section{4D GNN Mini-batch Training}
\label{sec:training}
We organize the total $G$ GPUs into a virtual 4D grid of size $G_d \times G_x
\times G_y \times G_z$.
The four dimensions serve complementary roles.
Data parallelism ($G_d$) replicates the training pipeline across independent 
groups, with each group processing a different mini-batch and synchronizing 
gradients through all-reduce.
Within each group, 3D PMM ($G_x \times G_y \times G_z$)
distributes the sparse and dense matrix operations across GPUs.
Every GPU also runs a communication-free sampling pipeline that constructs the 
mini-batch subgraph partitions from local data.

\subsection{Data Parallelism}
\label{subsec:data-parallel}

We partition the $G$ GPUs into $G_d$ groups of $G_x \times G_y \times G_z$ GPUs
each.
Each group holds a full copy of the model, distributed across GPUs via 3D
PMM (Section~\ref{subsec:3d-training}), and processes a distinct mini-batch 
at every training step.
The $G_d$ groups synchronize weight gradients through all-reduce after the 
backward pass, followed by the optimizer step.
Since each group trains on an independently sampled mini-batch, the effective 
batch size and aggregate throughput scale proportionally with $G_d$, while 
per-group computation and communication remain unchanged.

This design fundamentally differs from other baseline frameworks compared in 
our experiments (Section~\ref{sec:experiments}). In these systems, the graph 
is partitioned across all processes, and each process must fetch remote 
neighbors from other processes during sampling. In \plexusname, tensor 
parallelism (3D PMM) keeps the entire graph within each 
data-parallel group, eliminating all graph data movement. The only 
communication between data-parallel groups is gradient synchronization, which 
accounts for only a small fraction of per-epoch time as shown in our scaling 
experiments (Figure~\ref{fig:breakdown-scaling}).

\subsection{Distributed Sampling and Subgraph Construction}
\label{subsec:distributed-sampling}

Within a single data-parallel group of $G_x \times G_y \times G_z$ GPUs, each 
GPU holds a 2D shard of the adjacency matrix.
The central challenge is to construct a consistent $B \times B$ mini-batch 
subgraph across all GPUs without inter-device communication.
Uniform vertex sampling (Section~\ref{subsec:sampling-overview}) makes this 
possible. Because the sampled vertex set $\mathcal{S}$ depends only on a 
shared random seed and the graph size $N$, every GPU can independently derive 
the same $\mathcal{S}$ and extract its local portion of the induced subgraph.
Algorithm~\ref{alg:distributed-sampling} formalizes this per-GPU procedure in 
four phases, and Figure~\ref{fig:sampling} illustrates the distributed extraction.

\begin{figure*}[t]
  \centering
  \includegraphics[width=1\textwidth]{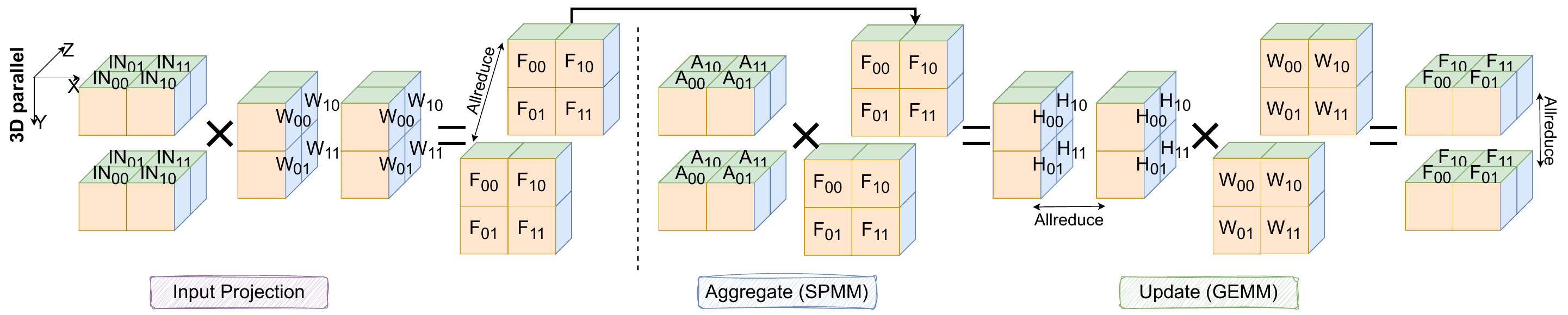}
  \caption{3D PMM forward pass in \plexusname with eight GPUs arranged
  in a $2{\times}2{\times}2$ grid ($X{\times}Y{\times}Z$). Left: the input
  projection multiplies the input feature shards (IN, on the $ZX$-plane) by
  weight shards ($W$, on the $XY$-plane) and an all-reduce along $Z$ produces 
  the projected features ($F$, on the $XY$-plane). Center: SpMM aggregation multiplies
  adjacency shards ($A$, on the $ZX$-plane) by $F$, followed by an all-reduce
  along $X$ to obtain the aggregated features ($H$). Right: the GEMM update
  multiplies $H$ by weight shards ($W$, on the $YX$-plane), and an all-reduce
  along $Y$ yields the layer output.}
  \label{fig:4d-training}
\end{figure*}

\begin{algorithm}[h]
\caption{Distributed Subgraph Construction (per GPU)}\label{alg:distributed-sampling}
\begin{algorithmic}[1]
\REQUIRE Local CSR shard $(\texttt{rp},\, \texttt{ci},\, \texttt{val})$ with row range $[R_0, R_1)$ and column range $[C_0, C_1)$, base seed $s$, step $t$, batch size $B$, graph size $N$
\STATE $\mathcal{S} \leftarrow \textsc{Sort}\bigl(\textsc{RandPerm}(N,\;\text{seed}{=}s{+}t)\,[{:}B]\bigr)$
\STATE $p \leftarrow (B{-}1)/(N{-}1)$ \COMMENT{inclusion probability, Eq.~\ref{eq:edge-scale}}
\STATE \COMMENT{\textbf{Phase 1: Locate local sample ranges}}
\STATE $\mathcal{S}_r \leftarrow \{v \in \mathcal{S} : R_0 \leq v < R_1\}$ via \textsc{BinarySearch}
\STATE $\mathcal{S}_c \leftarrow \{v \in \mathcal{S} : C_0 \leq v < C_1\}$ via \textsc{BinarySearch}
\STATE \COMMENT{\textbf{Phase 2: Vectorized CSR row extraction}}
\STATE $\mathbf{r} \leftarrow \texttt{rp}[\mathcal{S}_r{+}1] - \texttt{rp}[\mathcal{S}_r]$ \COMMENT{nnz per sampled row}
\STATE $\mathbf{P} \leftarrow \textsc{PrefixSum}(\mathbf{r})$
\STATE $\texttt{own} \leftarrow \textsc{SearchSorted}\bigl(\mathbf{P},\;\textsc{Arange}(\mathbf{P}[-1])\bigr)$
\STATE Gather $(\mathbf{i}_g,\, \mathbf{j}_g,\, \mathbf{v}_e)$ from CSR via \texttt{own}
\STATE \COMMENT{\textbf{Phase 3: Column filtering and compact remapping}}
\STATE $\texttt{mask} \leftarrow \textsc{Membership}(\mathbf{j}_g,\;\mathcal{S}_c)$ \COMMENT{binary search}
\STATE $(\mathbf{i}_g,\, \mathbf{j}_g,\, \mathbf{v}_e) \leftarrow$ apply \texttt{mask}
\STATE $(\mathbf{i}_c,\, \mathbf{j}_c) \leftarrow \textsc{TagRemap}(\mathbf{i}_g,\;\mathbf{j}_g,\;t)$ \COMMENT{$O(B)$ map update}
\STATE \COMMENT{\textbf{Phase 4: Rescale and assemble}}
\STATE $\mathbf{v}_e[k] \leftarrow \mathbf{v}_e[k] / p$ for all $k$ where $\mathbf{i}_g[k] \neq \mathbf{j}_g[k]$
\STATE $\tilde{A}^{\mathrm{loc}}, (\tilde{A}^T)^{\mathrm{loc}} \leftarrow \textsc{BuildCSR}(\mathbf{i}_c,\;\mathbf{j}_c,\;\mathbf{v}_e)$
\STATE $X_{\mathcal{S}} \leftarrow X[\mathcal{S}_r]$;\quad $Y_{\mathcal{S}} \leftarrow Y[\mathcal{S}_r]$
\end{algorithmic}
\end{algorithm}

We now walk through Algorithm~\ref{alg:distributed-sampling} in detail.
During process group initialization, all GPUs within a data-parallel group 
share a single random seed. At each training step, every GPU uses this seed 
together with the step index to derive the identical sorted sample $\mathcal{S}$ 
independently (Line~1).

\noindent\textbf{Local range identification (Lines 3--5).}
Each GPU owns a contiguous row range $[R_0, R_1)$ and column range $[C_0, C_1)$.
Since $\mathcal{S}$ is sorted, the local subsets $\mathcal{S}_r$ and $\mathcal{S}_c$ 
can be located via binary search in $O(\log B)$, avoiding a linear scan over 
the full sample.

\noindent\textbf{Vectorized CSR row extraction (Lines 6--10).}
To handle the irregular layout of sampled CSR rows, we vectorize the 
extraction with a prefix-sum-based indexing scheme.
We first read per-row nonzero counts from the CSR row pointer, compute a 
prefix sum to obtain a flat offset array, and use a sorted search to map each 
flat index back to its owning row.
All triples are then extracted through one coalesced gather in $O(\mathrm{nnz}_{\mathcal{S}})$
work with full GPU parallelism.

\noindent\textbf{Column filtering and compact remapping (Lines 11--15).}
We retain only edges whose target vertex belongs to $\mathcal{S}_c$ via 
binary-search membership testing, and remap the surviving global indices to a 
dense $[0, B)$ namespace.
Rather than zeroing an $N$-element map at every step, we maintain a persistent 
map tagged with the current step counter $t$, so only $O(B)$ entries require 
updating per iteration.

\noindent\textbf{Rescaling and assembly (Lines 15--18).}
For unbiased aggregation, we divide off-diagonal edge weights by 
$p$ (Eq.~\ref{eq:edge-scale}).
We then assemble both the forward and transpose CSR matrices in a single pass 
from the compact triples.

All four phases execute independently on each GPU with no inter-device 
communication, and together they produce a local shard of the mini-batch 
subgraph.

\subsection{3D Parallel Matrix Multiplication (3D PMM)}
\label{subsec:3d-training}

Given the mini-batch subgraph from distributed sampling, the next challenge is 
to distribute the forward and backward passes of each GNN layer across the
$G_x \times G_y \times G_z$ GPUs within a data-parallel group. We adapt Agarwal
et al.'s 3D PMM algorithm~\cite{agarwal-3d} to the
mixed sparse-dense computation of GCN layers.

\subsubsection{Matrix Distribution}

We distribute the matrices involved in each GNN layer across orthogonal planes 
of the 3D grid (Figure~\ref{fig:4d-training}).
In the following, we focus on the first layer. The distribution strategy for 
other layers is similar.
For the input projection, we shard the raw input features across the
$ZX$-plane and the weight matrix across the $YZ$-plane. The resulting
projected features $F$, sharded across the $XY$-plane, serve as input to the
SpMM aggregation in the first GCN layer.
For the first GCN layer, we shard the adjacency matrix $\tilde{A}_{\mathcal{S}}$ 
across the $ZX$-plane and replicate it along $Y$, and shard the weight matrix $W$ 
across the $XY$-plane and replicate it along $Z$.
This layout ensures that each GPU stores only a small fraction of each matrix and 
can perform local matrix multiplications with minimal communication overhead.
For the other layers and output projection, the sharding dimensions depend on 
the number of GCN layers. We refer to this cyclic reassignment as 
\emph{layer rotation} and describe it in Section~\ref{subsec:layer-rotation}.

\subsubsection{Parallel GCN Convolution}

\noindent\textbf{Aggregation (SpMM).}
Each GPU multiplies its local adjacency shard by its local matrix $F$.
Because $\tilde{A}_{\mathcal{S}}$ and $F$ reside on different planes, each 
local product is a partial sum that an all-reduce across the $X$-parallel 
group combines into the complete aggregated features:
\begin{equation}
  H_{\mathrm{agg}} = \mathrm{AllReduce}_X\!\bigl(\tilde{A}_{\mathcal{S}}^{\,\mathrm{local}} \cdot F^{\mathrm{local}}\bigr).
\label{eq:3d-spmm}
\end{equation}

\noindent\textbf{Combination (GEMM).}
Each GPU then multiplies the aggregated features by its local weight shard, and
an all-reduce across the $Y$-parallel group combines the partial outputs:
\begin{equation}
  H_{\mathrm{out}} = \mathrm{AllReduce}_Y\!\bigl(H_{\mathrm{agg}} \cdot W^{\mathrm{local}}\bigr).
\label{eq:3d-gemm}
\end{equation}

The backward pass (Eqs.~\ref{eq:gemm-bwd-w}--\ref{eq:spmm-bwd}) follows the 
same parallel structure, replacing $\tilde{A}_{\mathcal{S}}$ with its 
transpose and reversing the order of operations.

\subsubsection{Layer Rotation}
\label{subsec:layer-rotation}

After the first GCN layer, the output lives on the $ZX$-plane and serves as the
feature matrix $F$ for the next layer.
However, this layout differs from the $XY$-plane distribution that the first
layer assumed for $F$, so reusing the same $ZX$-plane adjacency shard would
produce an incompatible local multiplication.
To resolve this, we store a separate adjacency shard for each of the next two layers,
$\tilde{A}_{\mathcal{S}}^{(1)}$ on the $YZ$-plane and
$\tilde{A}_{\mathcal{S}}^{(2)}$ on the $XY$-plane, so that the adjacency layout
always aligns with the current feature distribution.
The third layer's output returns to the $XY$-plane, and the cycle repeats with
period three.
This scheme requires at most three adjacency shards per GPU and adds no
communication overhead~\cite{ranjan:sc2025}.

\subsubsection{Other Parallel Operators}
\label{subsec:other-operators}

\noindent\textbf{Linear layers.}
We parallelize the input projection ($X_{\mathrm{in}} W_{\mathrm{in}}$) and output
head ($X_{h,L} W_{\mathrm{out}}$) in the same way as the GEMM in
Eq.~\ref{eq:3d-gemm}.

\noindent\textbf{Parallel RMS normalization.}
Because features are sharded along the column dimension across GPUs, computing
the sum of squares requires an all-reduce across the group that holds different
feature columns:
\begin{equation}
  \mathrm{RMS}(x) = \sqrt{\frac{1}{d_h}\,\mathrm{AllReduce}\!\bigl(\|x^{\mathrm{local}}\|^2\bigr)}.
\label{eq:parallel-rmsnorm}
\end{equation}
We then apply the normalization and learnable scale parameter locally without
further communication.

\noindent\textbf{Residual connections.}
The residual add $X_{h,l} = X_{d,l} + X_{h,l-1}$ requires the layer input and 
output to share the same sharding layout.
However, layer rotation (Section~\ref{subsec:layer-rotation}) changes the 
distribution plane at each layer. For example, layer $l$'s output may reside on the 
$ZX$-plane while its input $X_{h,l-1}$ lives on the $XY$-plane.
To resolve this mismatch, we reshard the residual tensor before the addition.
The resharding communication can overlap with compute kernels in the forward 
pass, such as SpMM, GEMM, and RMSNorm, further hiding its latency.

\noindent\textbf{Element-wise operators.}
ReLU and dropout operate independently on each element and require no 
communication.

\section{Optimization}
\label{sec:optimization}
\begin{figure}[h]
    \centering
    \includegraphics[width=\columnwidth]{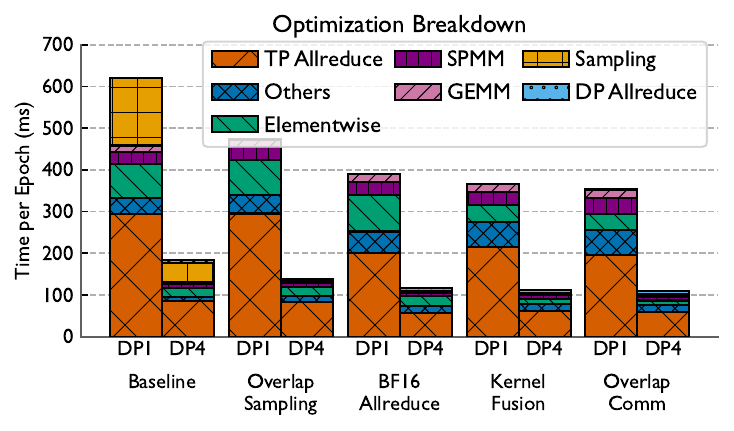}
    \caption{Breakdown of epoch times on ogbn-products
    with a $2{\times}2{\times}2$ grid per data-parallel group
    (DP1: 8 GPUs; DP4: 32 GPUs) as each optimization is applied
    cumulatively.}
    \label{fig:optimization}
\end{figure}

This section presents four optimizations that collectively reduce epoch time by
$1.75\times$ on eight GPUs and $1.66\times$ on 32 GPUs over the baseline 4D
pipeline.
We first profile the baseline on ogbn-products with a $2{\times}2{\times}2$
3D PMM grid (Figure~\ref{fig:optimization}, leftmost bars).
On eight GPUs (DP1), 3D PMM (tensor parallelism) all-reduce collectives 
account for 47\% of epoch time
and sampling accounts for 26\%; the remaining 27\% is split among element-wise
operations, SpMM, GEMM, and other overhead.
The breakdown at 32 GPUs (DP4) is similar, with an additional DP all-reduce
for gradient synchronization across data-parallel replicas.
The following subsections target these costs.

\subsection{Overlapping Sampling with Training}
\label{subsec:pipeline-opt}

Sampling and training stress \emph{complementary} hardware.
Sampling is bounded by GPU compute and memory bandwidth, while training at 
scale is dominated by collective communication.
We exploit this by \emph{prefetching} the next mini-batch.
Sampling and subgraph construction for step~$t{+}1$ run on a dedicated CUDA 
stream concurrently with the forward and backward passes of step~$t$.
The two streams are synchronized via a CUDA event before step~$t{+}1$ begins.
The same mechanism extends across epoch boundaries. The last step of epoch~$e$
prefetches the first mini-batch of epoch~$e{+}1$, so no step ever pays the full
sampling latency.
This overlap reduces epoch time by 24\% for both DP1 and DP4.

\subsection{Low-Precision Communication}
\label{subsec:mixed-precision}

With sampling off the critical path, all-reduce collectives become the dominant
bottleneck. Mixed-precision training is well established in deep 
learning~\cite{micikevicius2018mixed, kalamkar2019study}.
However, research shows that naively running full GNN layers in half precision 
degrades accuracy~\cite{gnn-half-precision}.

Inspired by this line of work, we apply reduced precision selectively to 
\emph{communication} rather than computation.
We cast FP32 partial sums to BF16 before the all-reduce and cast back 
afterward, but only for the collectives arising from 3D PMM.
For numerically sensitive reductions, such as all-reduce in parallel RMSNorm 
and logit reduction in parallel cross-entropy, we retain FP32 to preserve 
numerical stability.
All local computation in SpMM, GEMM, and element-wise operations remains in FP32.
This strategy halves the communication volume of the dominant collectives 
while avoiding precision loss in numerically sensitive operations.
We verified this approach across multiple runs on ogbn-products and Reddit 
datasets and found that the test-accuracy curves of BF16 communication are 
indistinguishable from full FP32 training. This approach further reduces 
epoch time by 17\% (DP1) and 16\% (DP4).

\subsection{Kernel Fusion}
\label{subsec:kernel-fusion}

With communication reduced, element-wise operators become a visible fraction of  
epoch time. In the baseline, each GNN layer applies RMSNorm, ReLU, and dropout 
as three separate CUDA kernels with redundant memory round-trips.
We use \texttt{torch.compile} to fuse these into a single kernel that eliminates
intermediate HBM transfers. Because the fused operations are purely element-wise, 
no change to the communication structure is required.
The compilation overhead is a one-time cost. 
This further reduces epoch time by 6\% (DP1) and 4\% (DP4).

\subsection{Overlapping Communication with Computation}
\label{subsec:overlap-comm}

In the backward pass, 3D PMM requires multiple all-reduces per layer on
orthogonal process groups ($X$-, $Y$-, and $Z$-groups). Since NCCL schedules
operations on different communicators independently, we overlap the
feature gradient all-reduce ($\nabla_{H_\mathrm{agg}}$) with the local
weight gradient computation ($\nabla_W$), and further overlap the weight
gradient synchronization with the feature gradient
communication. Similarly, within the linear layers, the input gradient
all-reduce ($\nabla_X$) and the weight gradient all-reduce ($\nabla_W$)
operate on orthogonal groups and thus run concurrently. This further reduces epoch time by 3\% (DP1) and 
2\% (DP4), bringing the cumulative speedup to $1.75\times$ (DP1) and 
$1.66\times$ (DP4) over the baseline.

\section{Experimental Setup}
\label{sec:experiments}
This section describes the experimental setup, including the target systems, 
graph datasets, and baseline frameworks.

\subsection{Systems and Environment}
\label{subsec:systems}

\noindent{\bf Perlmutter} is an HPE Cray EX system at NERSC.
We use its GPU nodes, each with four NVIDIA A100 GPUs.
Nodes are connected through HPE Slingshot-11 in a dragonfly topology, and four
Cassini NICs per node provide 100\,GB/s aggregate injection bandwidth.

\noindent{\bf Frontier} is an HPE Cray EX exascale system at OLCF.
Each compute node contains four AMD Instinct MI250X GPUs.
Each MI250X comprises two GPU dies (GCDs) with 64\,GB HBM2e per GCD, yielding
eight GCDs per node.
Frontier uses the same HPE Slingshot-11 interconnect, with four NICs per node
and 100\,GB/s aggregate bandwidth.

\noindent{\bf Tuolumne} is an HPE Cray EX system at LLNL.
Each compute node contains four AMD MI300A APUs, each integrating GPU and CPU
dies on a single package with 128\,GB unified HBM3 memory.
Nodes are connected through HPE Slingshot-11, with four NICs per node providing
100\,GB/s aggregate injection bandwidth.

\subsection{Baseline Frameworks}
\label{subsec:baselines}

We compare \plexusname against four distributed GNN training systems that
represent distinct parallelization and sampling strategies.
For all baseline frameworks, we combine the configurations recommended in 
their papers with our own hyperparameter sweeps to select the best-performing 
settings. \textbf{BNS-GCN}~\cite{bns-gcn} performs full-graph training and
reduces cross-partition communication by sampling boundary vertices.
We use a sampling ratio of 0.1, which the authors report as optimal.
\textbf{DistDGL}~\cite{distdgl} extends the Deep Graph Library~\cite{dgl}
to multi-node clusters by partitioning the graph and storing vertex features 
in a distributed key-value store, with each worker running neighbor sampling 
and training locally.
\textbf{MassiveGNN}~\cite{massivegnn} builds on DistDGL with optimized
distributed feature fetching for GraphSAGE-style neighbor sampling.
\textbf{SALIENT++}~\cite{kaler2023communication} accelerates
GraphSAGE-style neighbor sampling on CPUs and uses feature caching to reduce
remote memory accesses.

\subsection{Datasets and Evaluation Methodology}
\label{subsec:datasets}

We evaluate \plexusname on five graph datasets that span different domains and
scales.
\textbf{ogbn-products}~\cite{ogb} (product category prediction) and
\textbf{Reddit}~\cite{graphsage} (community classification) are widely used
benchmarks for distributed GNN training. We use them to validate model accuracy
and compare end-to-end training performance against baseline frameworks.
We additionally select three larger datasets to demonstrate the scaling
capability of \plexusname.
\textbf{Isolate-3-8M}~\cite{hipmcl} is a subgraph extracted from a protein
similarity network with 3.8\,M vertices.
\textbf{Products-14M}~\cite{ni-etal-2019-justifying} is a larger Amazon product
network with 14\,M vertices and 115\,M edges.
\textbf{ogbn-papers100M}~\cite{ogb} is a citation network with 111\,M vertices
and 1.6\,B edges.
For datasets that do not include node features (Isolate-3-8M and Products-14M),
we generate random input features with dimension~128 and assign 32 synthetic
classes proportional to vertex degree.
Since these two datasets are used to measure scaling efficiency rather
than model accuracy, the synthetic features do not affect the validity of the
results.

\paragraph{Cross-framework comparison.}
Unlike images or text, graph data is not independently and identically 
distributed (i.i.d.).
Each mini-batch subgraph captures only a partial view of the global graph 
structure, and its information content depends on the sampling algorithm. 
Different sampling algorithms produce mini-batches with different structural 
coverage and statistical properties. Aggregating the mini-batches within a
single epoch does not recover the full graph, and the gap varies across
methods. Consequently, \emph{time per epoch} is only a meaningful metric when
comparing runs that use the same sampling algorithm.
An ``epoch'' under one sampling strategy is therefore not comparable
to an epoch under another, so comparing time per epoch across
frameworks is not meaningful. We instead adopt \emph{end-to-end training time} 
to a target test accuracy as the primary metric for cross-framework 
comparison, which reflects the real cost a practitioner faces.
We report total wall-clock time to reach a target test accuracy of 95\% on Reddit
and 79\% on ogbn-products, which reflect the converged GCN accuracy reported
across multiple prior works~\cite{graphsaint,graphsage,bns-gcn,ogb}.
The reported times in this paper include only training time.

\paragraph{Scaling experiments.}
To understand the scaling behavior of \plexusname, we evaluate it on all five
datasets and report time per epoch as the performance metric.

\section{Results}
\label{sec:results}
This section presents the results of our end-to-end training and scaling
experiments across five graph datasets, and compares \plexusname with four
baseline frameworks.

\begin{figure*}[t]
    \centering
    \includegraphics[width=\textwidth]{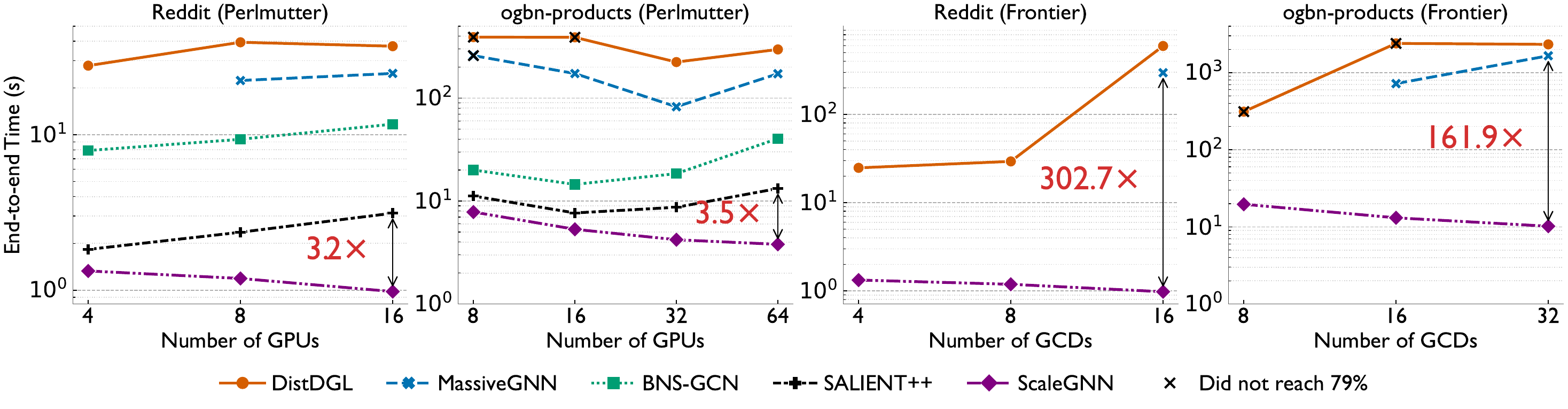}
    \caption{End-to-end training time to reach target test accuracy on Perlmutter and Frontier (log scale).
    Lower is better. Points marked with $\times$ did not reach the target accuracy. 
    We report the time to their best observed accuracy.
    On Perlmutter: DistDGL 78.57\% at 8 GPUs, 78.95\% at 16 GPUs; MassiveGNN 78.59\% at 8 GPUs.
    On Frontier: DistDGL on ogbn-products 77.45\% at 8 GCDs, 78.9\% at 16 GCDs.
    Annotations compare \plexusname to SALIENT++ (Perlmutter) and MassiveGNN (Frontier) at the largest GPU/GCD count for each dataset.}
    \label{fig:e2e}
\end{figure*}  

\subsection{Sampling Accuracy}
\label{subsec:sampling-accuracy}

We first verify that our uniform vertex
sampling with unbiased edge rescaling does not sacrifice model quality.
We compare against two representative sampling strategies.
GraphSAINT~\cite{graphsaint} is a widely used subgraph sampling method. We
compare with its node sampling variant, which is closest to our approach.
GraphSAGE~\cite{graphsage} is a neighbor sampling method adopted by several
baseline systems including DistDGL, MassiveGNN, and SALIENT++.

\begin{table}[h]
\centering
\caption{Test accuracy (\%) comparison}
\label{tab:sampling-accuracy}
\small
\begin{tabular}{lcc}
\hline
\textbf{System} & \textbf{Reddit} & \textbf{ogbn-products} \\
\hline
GraphSAINT (node)      & 96.2 & 80.2 \\
GraphSAGE       & 95.4 & 79.6 \\
\rowcolor{blue!10} \plexusname     & \textbf{96.3} & \textbf{81.3} \\
\hline
\end{tabular}
\end{table}

Table~\ref{tab:sampling-accuracy} reports the best test accuracy achieved by 
each algorithm.
On both datasets, \plexusname matches or slightly exceeds the accuracy of
GraphSAINT node sampling and GraphSAGE neighbor sampling.
On ogbn-products, \plexusname achieves 81.3\%, outperforming GraphSAINT (80.2\%)
by 1.1 percentage points and GraphSAGE (79.6\%) by 1.7 percentage points.
This confirms that uniform vertex sampling with unbiased edge rescaling
(Eq.~\ref{eq:rescaled-adj}) preserves model quality.

The following sections demonstrate the strong performance and scalability 
enabled by communication-free sampling algorithm, through end-to-end 
training and scaling experiments.

\subsection{End-to-End Performance}
\label{subsec:e2e-perf}

Figure~\ref{fig:e2e} reports end-to-end training time to target accuracy on
Reddit and ogbn-products on both Perlmutter and Frontier.
On Frontier, BNS-GCN and SALIENT++ do not provide ROCm support, so we compare
\plexusname only against DistDGL and MassiveGNN.
Additionally, MassiveGNN only supports multi-node execution, so its smallest 
configuration is 8 GPUs on Perlmutter and 16 GCDs on Frontier.

On Reddit (Figure~\ref{fig:e2e}, first panel), \plexusname trains to target accuracy in
1.33\,s at 4 GPUs and 0.98\,s at 16 GPUs, consistently outperforming all baselines.
SALIENT++ starts at 1.83\,s on 4 GPUs but slows to 3.13\,s on 16 GPUs.
BNS-GCN shows a similar trend, rising from 7.92\,s to 11.7\,s.
DistDGL and MassiveGNN are over an order of magnitude slower.
At 16 GPUs, \plexusname achieves a $3.2\times$ speedup over SALIENT++ and
$11.9\times$ over BNS-GCN.
On ogbn-products (Figure~\ref{fig:e2e}, second panel), \plexusname reaches the target
accuracy in 7.83\,s at 8 GPUs, compared with 11.19\,s for SALIENT++ and
20.02\,s for BNS-GCN, yielding speedups of $1.4\times$ and $2.6\times$.
At 64 GPUs, \plexusname finishes in 3.80\,s, while SALIENT++ requires 13.25\,s
and BNS-GCN 40.46\,s, yielding speedups of $3.5\times$ and $10.6\times$,
respectively.

On Frontier, \plexusname shows similar advantages over the two available
baselines.
On Reddit, \plexusname reaches the target accuracy in 0.98\,s at 16 GCDs,
while DistDGL requires 596.09\,s and MassiveGNN 296.69\,s.
On ogbn-products, \plexusname finishes in 10.2\,s at 32 GCDs, compared with
1651.73\,s for MassiveGNN and 2321.34\,s for DistDGL, yielding speedups of
$162\times$ and $228\times$.
DistDGL on ogbn-products at 8 and 16 GCDs did not reach the 79\% target
accuracy.

For frameworks that fail to reduce end-to-end training time with more GPUs,
we observe that increasing data parallelism raises the number of epochs
needed to reach the target accuracy.
Additionally, these baselines do not proportionally reduce epoch time with
additional GPUs. The extra epochs outweigh the throughput gain, resulting in
longer end-to-end time.

\begin{table}[h]
\centering
\caption{Time per evaluation round}
\label{tab:eval-time}
\resizebox{\columnwidth}{!}{
\begin{tabular}{lcc}
\hline
\textbf{System} & \textbf{Reddit (4 GPUs)} & \textbf{ogbn-products (8 GPUs)} \\
\hline
DistDGL / MassiveGNN & 12.50\,s & 20.82\,s \\
SALIENT++            & 1.13\,s  & 10.12\,s \\
BNS-GCN             & 1.79\,s  & 6.89\,s  \\
\rowcolor{blue!10} \plexusname          & \textbf{0.05\,s}  & \textbf{0.19\,s}  \\
\hline
\end{tabular}
}
\end{table}

Although the training times in Figure~\ref{fig:e2e} exclude evaluation, evaluation
costs can significantly impact end-to-end wall-clock time in practice.
Table~\ref{tab:eval-time} compares the time of each evaluation round.
Because \plexusname distributes the full graph and model across GPUs via 3D 
PMM, it performs full-graph evaluation with a single distributed forward pass 
and no sampling overhead. 
In contrast, all four baselines use data-parallel training with a fully 
replicated model, and their evaluation pipelines cannot leverage multiple 
GPUs effectively.
SALIENT++ and DistDGL/MassiveGNN rely on neighbor sampling during evaluation to
fit within GPU memory, so evaluation still requires the same multi-hop 
sampling and remote feature fetching pipeline used during training, incurring 
substantial overhead.
BNS-GCN avoids fanout-based sampling but falls back to single-process 
full-graph inference on the CPU, so it cannot leverage GPU acceleration 
or distribute the evaluation workload at all.
On ogbn-products at eight GPUs, \plexusname evaluates in 0.19\,s per round,
$36\times$ faster than BNS-GCN (6.89\,s), $54\times$ faster than SALIENT++
(10.12\,s), and $111\times$ faster than DistDGL/MassiveGNN (20.82\,s).
On Reddit at four GPUs, \plexusname finishes in 0.05\,s
per round, achieving $36\times$, $23\times$, and $250\times$ speedups over
BNS-GCN, SALIENT++, and DistDGL/MassiveGNN, respectively.

\begin{figure*}[t]
    \centering
    \includegraphics[width=\textwidth]{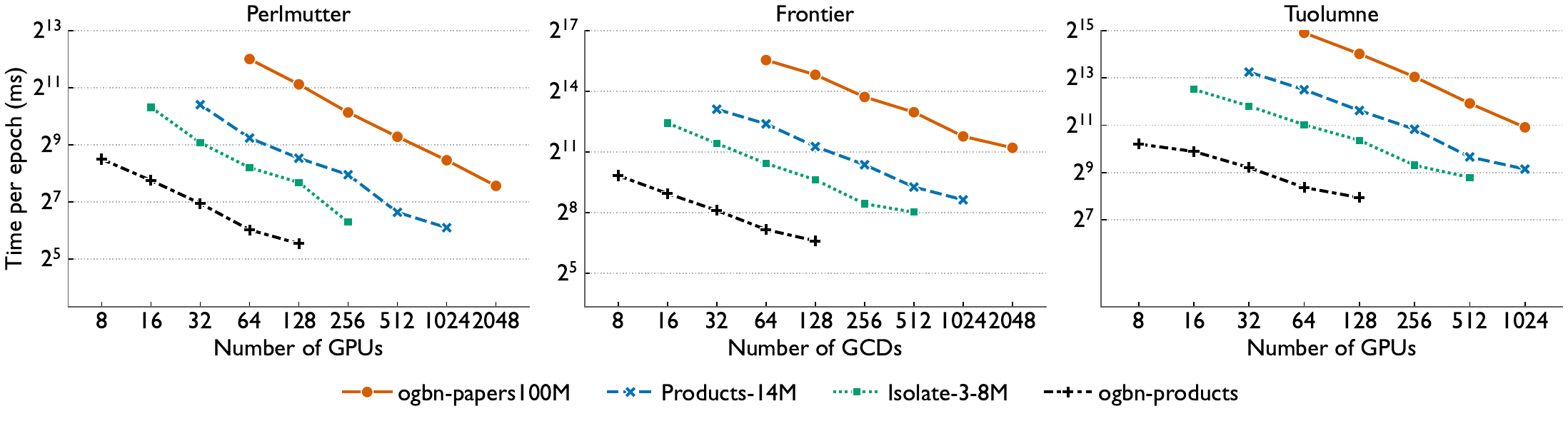}
    \caption{Strong scaling on Perlmutter (left), Frontier (center), and Tuolumne (right). Each
    curve starts at the smallest 3D PMM configuration ($G_d{=}1$)
    and scales out by increasing data-parallel replicas $G_d$.}
    \label{fig:scaling}
\end{figure*}  

\subsection{Scaling Results}
\label{subsec:scaling}

\begin{figure}[h]
    \centering
    \includegraphics[width=\columnwidth]{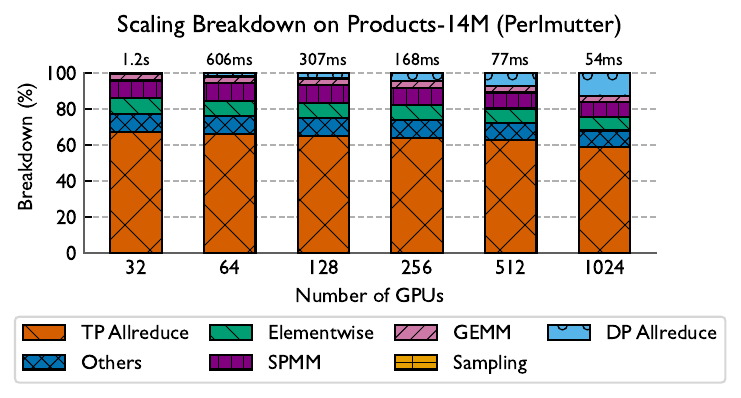}
    \caption{Epoch time breakdown on Products-14M on Perlmutter.}
    \label{fig:breakdown-scaling}
\end{figure}

Figure~\ref{fig:scaling} reports epoch time on Perlmutter (left), Frontier
(center), and Tuolumne (right).
Following prior work on 3D PMM~\cite{agarwal-3d,
ranjan:sc2025,singh:ipdps2022,singh:sc2024}, we choose
$G_x \times G_y \times G_z$ to be as close to a cube as possible, as this is
the most efficient configuration.
For each dataset, the leftmost point in the scaling curve corresponds to
$G_d = 1$, i.e., pure tensor parallelism with no data-parallel replication.
We then scale to larger GPU counts by increasing $G_d$ while keeping the 3D
PMM configuration fixed.

On Perlmutter, \plexusname demonstrates consistent strong scaling across all
datasets. On ogbn-papers100M, \plexusname scales from 64 to 2048 GPUs with
a $21.7\times$ speedup, reducing epoch time from 4095\,ms to 189\,ms.
On Products-14M, \plexusname achieves $19.8\times$ speedup from 32 to
1024 GPUs.
On Isolate-3-8M, scaling from 16 to 256 GPUs yields a $16.2\times$ speedup.
On ogbn-products, the smallest dataset, \plexusname scales from 8 to 128 GPUs
with a $7.8\times$ speedup.

On Frontier, \plexusname shows similar scaling trends.
On Products-14M, \plexusname achieves $22.4\times$ speedup from 32 to
1024 GCDs, reducing epoch time from 8809\,ms to 394\,ms.
On Isolate-3-8M, scaling from 16 to 512 GCDs yields a $21.2\times$ speedup.
On ogbn-papers100M, \plexusname scales from 64 to 2048 GCDs with a
$20.3\times$ speedup.
Epoch times on Frontier are higher than on Perlmutter due to differences in
GPU architectures and communication libraries. Prior work~\cite{pccl} has also
shown that RCCL achieves lower communication throughput than NCCL at scale.
Despite this, the scaling efficiency remains comparable across both systems.

On Tuolumne, \plexusname achieves $17.2\times$ speedup on Products-14M from 32
to 1024 GPUs, reducing epoch time from 9710\,ms to 566\,ms.
On ogbn-papers100M, scaling from 64 to 1024 GPUs yields a $15.9\times$ speedup.
On Isolate-3-8M, scaling from 16 to 512 GPUs yields a $13.2\times$ speedup.

Figure~\ref{fig:breakdown-scaling} breaks down epoch time by component on
Products-14M as we increase $G_d$.
At $G_d = 1$, the all-reduce for data-parallel gradient synchronization is 
absent, and epoch time is dominated by the tensor-parallel (3D PMM) 
collectives and compute kernels.
As $G_d$ grows, the data-parallel all-reduce cost rises from negligible to an
increasing fraction of epoch time, reflecting the growing
communication volume of gradient synchronization across more replica groups.
Meanwhile, the time spent on 3D PMM operations and sampling remains
roughly constant, confirming that data parallelism scales the training pipeline
without inflating per-group work.

\section{Conclusion}
\label{sec:conclusion}
We present \plexusname, a 4D parallel framework for scalable mini-batch GNN
training that unifies distributed communication-free sampling, 3D parallel matrix multiplication,
and data parallelism. \plexusname extends 3D PMM to GNN layers by 
distributing both graph data and model weights across a 3D 
processor grid, enabling training on graphs and models that exceed 
single-GPU memory. We introduce a uniform vertex sampling strategy that 
enables communication-free distributed sampling. We further explore 
several optimization strategies to reduce training time.

Experiments on five graph datasets show that \plexusname's sampling strategy
matches or exceeds the accuracy of GraphSAINT and GraphSAGE.
On Perlmutter, \plexusname achieves $3.5\times$ end-to-end training
speedup over the SOTA baseline on ogbn-products.
Strong scaling experiments demonstrate efficient scaling to 2048 GPUs on
Perlmutter, 2048 GCDs on Frontier, and 1024 GPUs on Tuolumne.

\section*{Acknowledgment}
This material is based upon work supported by the National Science Foundation
under Grant No.~2047120.  This research used resources of the National Energy
Research Scientific Computing Center (NERSC), a U.S.~Department of Energy (DOE)
Office of Science User Facility, operated under Contract No.~DE-AC02-05CH11231
using NERSC awards DDR-ERCAP0034262 and ALCC-ERCAP0034775, and that of the Oak Ridge Leadership
Computing Facility at the Oak Ridge National Laboratory, which is supported by
the Office of Science of the U.S. DOE under Contract No. DE-AC05-00OR22725.
This work was performed under the auspices of the U.S.~Department of Energy
(DOE) by Lawrence Livermore National Laboratory under Contract
DE-AC52-07NA27344 (LLNL-CONF-XXX).

\bibliographystyle{IEEEtran}
\bibliography{../bib/cite,../bib/pssg}

\end{document}